\documentclass{article}
\usepackage{arxiv}

\usepackage[utf8]{inputenc} 
\usepackage[T1]{fontenc}    
\usepackage{hyperref}       
\usepackage{url}            
\usepackage{datetime}       
\usepackage{booktabs}       
\usepackage{amsfonts}       
\usepackage{nicefrac}       
\usepackage{microtype}      
\usepackage{graphicx}
\usepackage{natbib}
\usepackage{doi}
\usepackage{xcolor}

\hypersetup{colorlinks = true,
linkcolor = purple,
urlcolor  = blue,
citecolor = cyan,
anchorcolor = black}

\title{Deep Learning for GWP Prediction: A Framework Using PCA, Quantile Transformation, and Ensemble Modeling}

\date{}

\makeatletter
\let\@fnsymbol\@arabic
\makeatother

\author{\href{https://orcid.org/0000-0001-5318-5983}{\includegraphics[scale=0.06]{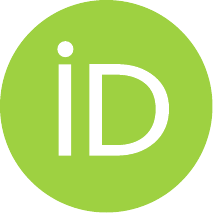}}\hspace{1mm}Navin Rajapriya\footnotemark[1]\\
AIZOTH Inc., 2nd Floor, Daiwa Roynet Hotel Tsukuba Bldg, 1-5-7 Azuma, Tsukuba, IBARAKI 305-0031, JAPAN\\\AND
\href{https://orcid.org/0000-0001-6862-802X}{\includegraphics[scale=0.06]{orcid.pdf}}\hspace{1mm}Kotaro Kawajiri\\
AIZOTH Inc., 2nd Floor, Daiwa Roynet Hotel Tsukuba Bldg, 1-5-7 Azuma, Tsukuba, IBARAKI 305-0031, JAPAN\\}

\renewcommand{\headeright}{}
\renewcommand{\undertitle}{
}
\renewcommand{\shorttitle}{}

\hypersetup{
pdftitle={\@title},
pdfsubject={},
pdfauthor={\@author},
pdfkeywords={},
addtopdfcreator={Written in Curvenote}
}

\begin{document}
\maketitle
\footnotetext[1]{Correspondence to: navin.rajapriya@aizoth.com}

\begin{abstract}
Developing environmentally sustainable refrigerants is critical for mitigating the impact of anthropogenic greenhouse gases on global warming. This study presents a predictive modeling framework to estimate the 100-year global warming potential (GWP100) of single-component refrigerants using a fully connected neural network implemented on the Multi-Sigma platform. Molecular descriptors from RDKit, Mordred, and alvaDesc were utilized to capture various chemical features. The RDKit-based model achieved the best performance, with a Root Mean Square Error (RMSE) of 481.9 and an R\textsuperscript{2} of 0.918, demonstrating superior predictive accuracy and generalizability.

Dimensionality reduction through Principal Component Analysis (PCA) and quantile transformation were applied to address the high-dimensional and skewed nature of the dataset, enhancing model stability and performance. Factor analysis identified vital molecular features, including molecular weight, lipophilicity, and functional groups, such as nitriles and allylic oxides, as significant contributors to GWP values. These insights provide actionable guidance for designing environmentally sustainable refrigerants.

Integrating RDKit descriptors with Multi-Sigma's framework---including PCA, quantile transformation, and neural networks---offers a scalable solution for rapid virtual screening of low-GWP refrigerants. This approach can potentially accelerate the identification of eco-friendly alternatives, directly contributing to climate mitigation by enabling the design of next-generation refrigerants aligned with global sustainability objectives.
\end{abstract}


global warming potential (GWP), refrigerants, deep learning, molecular descriptors

\section{Introduction}

Global warming is a pressing environmental issue, driven primarily by anthropogenic greenhouse gases. Among these, refrigerants---used extensively in cooling applications such as air conditioning, refrigeration, and industrial processes---significantly contribute to global warming due to their high Global Warming Potential (GWP). For example, commonly used refrigerants such as R-134a have a GWP over 1,400 times that of carbon dioxide, contributing extensively to climate change [1]. In 2022, hydrofluorocarbons (HFCs) alone accounted for about 11\% of the overall greenhouse effect caused by human activities [2]. This makes addressing refrigerants critical in reducing overall emissions and achieving climate targets. The Kigali Amendment to the Montreal Protocol has established international regulations to reduce the production and consumption of high-GWP HFCs, reinforcing the urgent need to identify and develop alternative refrigerants with significantly lower GWP (\textless 100) [3][4]. Achieving this goal necessitates innovative, data-driven approaches to design refrigerants that balance performance requirements with environmental sustainability [5][6].

Designing low-GWP refrigerants involves navigating complex structure-property relationships. Traditional group contribution methods, though widely used, assume linear correlations between molecular structure and properties, limiting their ability to capture the non-linear effects and higher-order interactions that define GWP [7][8]. This limitation becomes more pronounced in complex molecules where 3D spatial information and stereochemical features are critical. Addressing these challenges requires advanced predictive tools like deep learning, which excel at uncovering intricate relationships between molecular features and environmental properties. However, the dimension of molecular descriptor packages, particularly when combining 2D and 3D descriptors, makes direct modeling of neural networks challenging despite its effectiveness for non-linear correlations [9]. Generally, feature selection, such as mutual information-permutation importance (MI-PI) and recursive feature elimination (RFE), is used to filter out low informative molecular descriptors to improve prediction accuracy [6][10][11]. However, feature selection can lead to model overfitting and positive bias, making it crucial to balance between reducing feature dimensionality for computational efficiency and retaining the critical information necessary for accurate model prediction.

The 100-year GWP values of refrigerants and greenhouse gases (GHGs) reported in the IPCC AR6 exhibit a positive skew, indicating that most data points are concentrated towards the lower end of the GWP scale. In previous studies, logarithmic transformation has often been employed to preprocess GWP data, as it effectively addresses positive skewness and compresses the dynamic range of highly skewed datasets [12][13]. However, while effective at reducing skew, the log transformation may still struggle with distributions that contain many extreme outliers or minimal values close to zero. An alternative preprocessing approach, such as quantile transformation, not only addresses skewness but also transforms the data into a uniform distribution, ensuring that all values are equally represented [14]. This can provide a more balanced approach for training machine learning models, improving the stability and performance of models by making the distribution of the target variable less dependent on the presence of extreme values.

This study hypothesizes that a deep learning framework using an ensemble of fully connected neural networks, trained with quantile transformation and dimension reduction strategies such as principal component analysis (PCA), can accurately predict GWP while maintaining generalizability to new compounds. By comparing different molecular descriptor packages---RDKit, Mordred, and alvaDesc---we aim to determine the most effective set of features to model GWP. RDKit descriptors were chosen for their computational efficiency in representing 2D features, Mordred for capturing a combination of 2D and 3D molecular characteristics, and alvaDesc for providing detailed 3D descriptors, including stereochemical information. Unlike traditional feature selection, we employed PCA to reduce dimensionality, thereby preserving as much of the feature space as possible while eliminating redundancy [15].

By advancing a deep learning-based prediction model that effectively generalizes to a diverse set of refrigerants, our study represents a significant step forward in predictive modeling for climate mitigation. This framework not only accelerates the identification of low-GWP refrigerants, supporting international initiatives like the Kigali Amendment but also provides a scalable solution for virtual screening across a broad library of chemical compounds. Integrating PCA and quantile transformation ensures a balanced trade-off between complexity, accuracy, and generalizability, enhancing the framework's suitability for large-scale applications.

\section{Methodology}

The GWP 100 dataset used in this study comprised 207 refrigerants and greenhouse gases (GHGs) from the IPCC AR6 dataset. The compounds included a wide range of chemical classes, such as hydrocarbons (HC), chlorofluorocarbons (CFC), hydrochlorofluorocarbons (HCFC), hydrofluorocarbons (HFC), hydrofluoroolefins (HFO), chlorocarbons (CC), and halogenated hydrocarbons (HCC). This diversity ensured that a broad spectrum of functional group variations was captured, including alkanes, alkenes, halides, and others, making the dataset representative of real-world refrigerant diversity. The overview of the deep learning framework employed in this study is illustrated in Figure 1.

\begin{figure}[!htbp]
\centering
\includegraphics[width=0.7\linewidth]{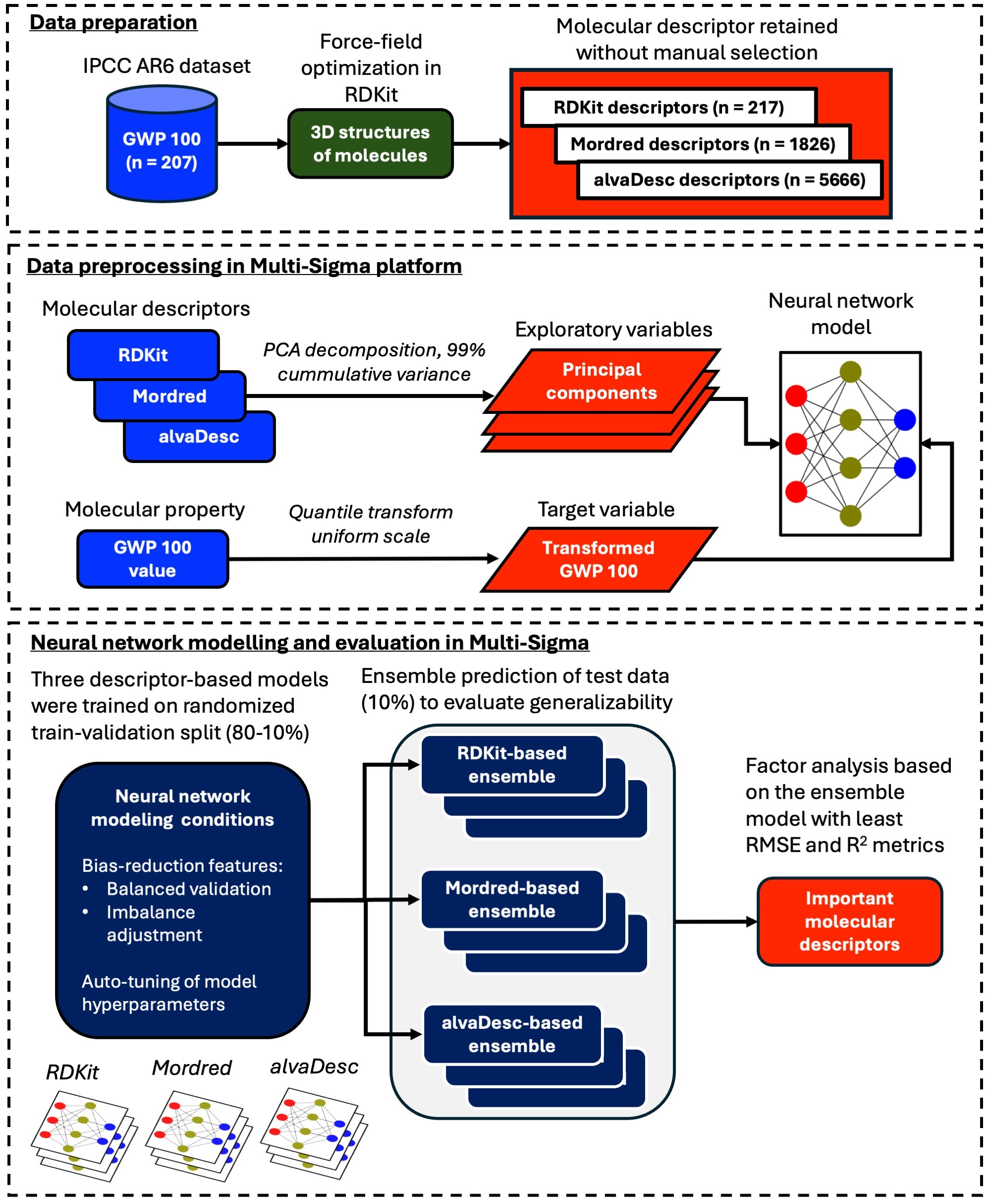}
\caption[]{Framework for predicting GWP values using molecular descriptors, PCA, quantile transformation, and neural networks, implemented on the Multi-Sigma platform}
\label{Um2Qt8Q6k8}
\end{figure}

\subsection{Data Preparation}

The 3D structures of the molecules were processed in RDKit using force-field optimization. Molecular descriptors were generated using three different packages: RDKit, Mordred, and alvaDesc. The choice of descriptor packages---RDKit, Mordred, and alvaDesc---was motivated by their complementary characteristics. RDKit descriptors (217 features) primarily include 1D and 2D features representing atom connectivity and simple molecular characteristics, which are computationally efficient. Mordred descriptors (1826 features) extend this representation to include a mix of 2D and 3D features, providing additional complexity. AlvaDesc descriptors (5666 features) emphasize detailed 3D features, capturing intricate spatial information. Three datasets were created using the descriptor packages to develop their corresponding neural network models.

The datasets were split into training, validation, and test sets using an 80-10-10 ratio. The test set (10\%) was common for the three descriptor-based datasets, while the training and the validation set were randomized during the modeling phase to build an ensemble prediction model.

\subsection{Data Preprocessing}

Given the relatively small dataset size of 207 samples, PCA was preferred over non-linear methods like autoencoders. Autoencoders typically require larger datasets to effectively learn non-linear relationships without overfitting, whereas PCA is computationally efficient and effectively captures approximately 99\% of data variance, reducing dimensionality while minimizing overfitting risks. We did not employ feature selection to avoid bias during model training and to create a generalizable model that can be used for direct virtual screening with minimal feature engineering.

To address the skewness in the GWP values (Figure 2), quantile transformation was applied to achieve a more uniform distribution, which is essential for stable model training. The transformed values were used for model training, and predictions were inversely transformed to interpret them on the original GWP scale. This step aimed to balance efficient learning with practical interpretability.

\begin{figure}[!htbp]
\centering
\includegraphics[width=0.8\linewidth]{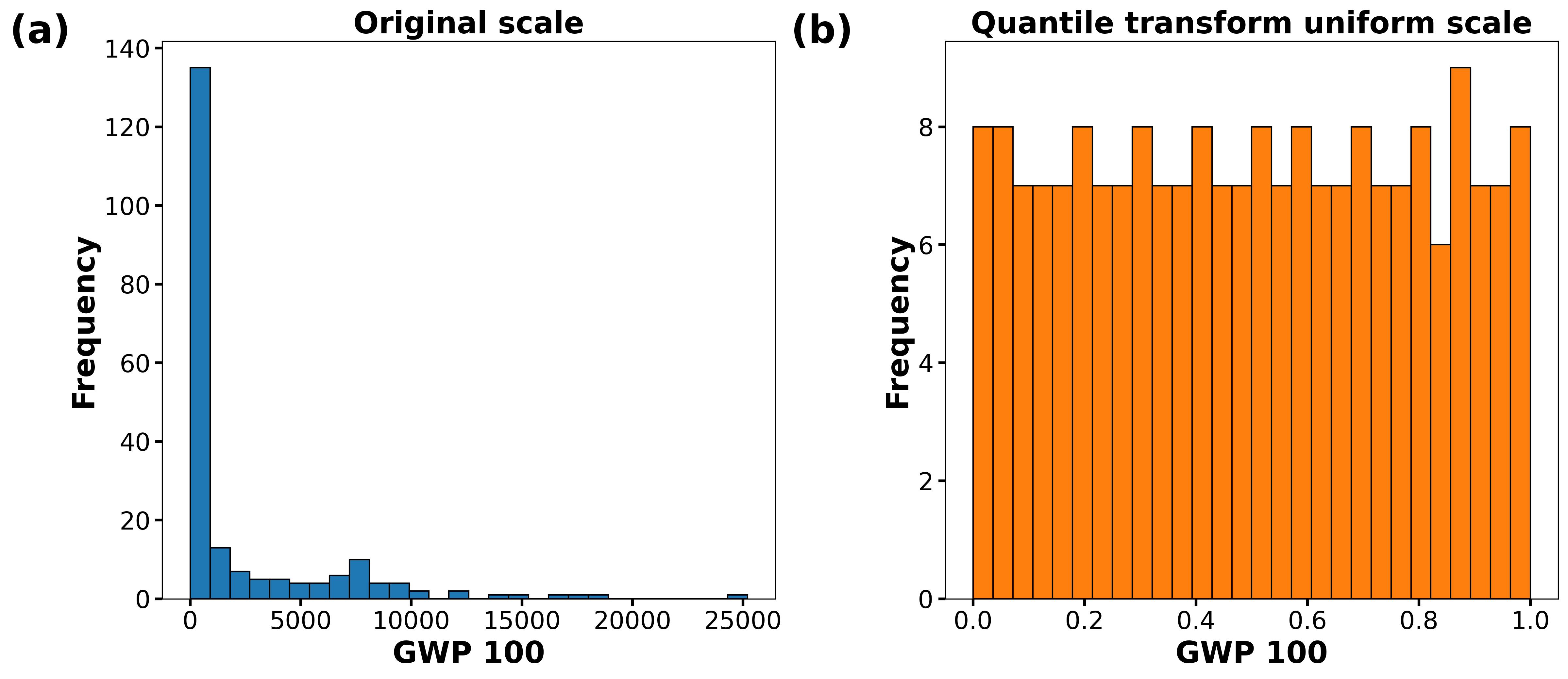}
\caption[]{Distribution of GWP100 values before and after quantile transformation: (a) Original scale (b) Quantile transform uniform scale}
\label{PZz8hKR9fj}
\end{figure}

\subsection{Neural Network Modeling and Evaluation}

A fully connected neural network architecture was employed for predicting GWP, utilizing the Multi-Sigma platform---a user-friendly deep learning tool with a graphical interface. Multi-Sigma's auto-tuning feature was leveraged to optimize hyperparameters, minimizing prediction error and mitigating overfitting during training. This automated process iteratively evaluates validation errors for various hyperparameter configurations and eliminates those leading to higher RMSE. The auto-tuning procedure optimized key network parameters, including the number of hidden layers, neurons per layer, activation functions, epochs, and batch sizes, ensuring robust and efficient model performance. Linear activation function was selected for the output layer, and the Adam optimizer with an initial learning rate of 0.001 was used for training. Two of Multi-Sigma's bias reduction features, leveraging balanced validation data sampling and imbalance data adjustment, were employed during the model training stage. Balanced validation sampling ensures that data distribution in the validation set remains consistent during sampling, preserving proportional representation across the dataset. Imbalance adjustment addresses underrepresented instances by up-sampling based on each feature and the target variable, improving the model's generalizability and ensuring robust prediction performance across diverse data subsets.

Up to ten prediction models were created for each descriptor package involving random sampling while maintaining the aforementioned constraints. For each descriptor package, top three models exhibiting the lowest root mean square error (RMSE) on the validation set were ensembled together. These ensemble models were verified for their robustness against the test dataset. The ensemble approach reduces the impact of outliers or biases that might arise from individual model variance [16][17]. The prediction accuracy of ensembled models for each descriptor package was evaluated using RMSE and R\textsuperscript{2} scores to identify the best ensemble for predicting GWP.

To evaluate the contribution of each PC to GWP prediction, factor analysis was conducted on the best-performing ensemble model, selected based on its lowest RMSE and highest R\textsuperscript{2} score. This analysis leveraged Multi-Sigma's permutation sensitivity-based feature importance approach to quantify the positive and negative impacts of each PC on GWP values. By identifying the most influential PCs, the analysis provided key insights into the critical features driving accurate predictions, improving model interpretability, and guiding future screening strategies. Furthermore, the original loadings of influential PCs were examined to understand the contributions of specific molecular descriptors and their broader implications for GWP reduction.

\section{Results and Discussion}

\subsection{Quantile Transformation and Dimensionality Reduction for Accurate GWP Prediction}

PCA was employed to reduce the number of features while retaining 99\% of the cumulative variance, ensuring the preservation of the most critical information and reducing the complexity of the molecular descriptors. Figure 3 illustrates the cumulative explained variance captured by the principal components for each descriptor package. The elbow point, typically used to determine an optimal number of components, was observed at around 90\% of the explained variance. However, to ensure high fidelity and capture as much complexity of the data as possible, a 99\% variance retention threshold was selected. This resulted in 48 PCs for RDKit, 73 for Mordred, and 99 for AlvaDesc.

The number of principal components required to retain 99\% variance varied significantly across descriptor packages. This variation can be explained by the differences in complexity and the nature of the molecular information within each descriptor package.

RDKit primarily provides dominantly 2D molecular descriptors, which are simpler and less redundant compared to the more complex 3D spatial features generated by Mordred and AlvaDesc. This explains why fewer principal components (48 PCs) were required to retain 99\% variance for RDKit, whereas Mordred and alvaDesc required more PCs (73 and 99, respectively).

Higher-order descriptors like those from alvaDesc include substantial redundancy due to the overlap of 3D geometric information. The increased redundancy requires more principal components to sufficiently represent the original dataset, whereas RDKit, with fewer overlapping features, captured the variance with fewer components.

\begin{figure}[!htbp]
\centering
\includegraphics[width=1\linewidth]{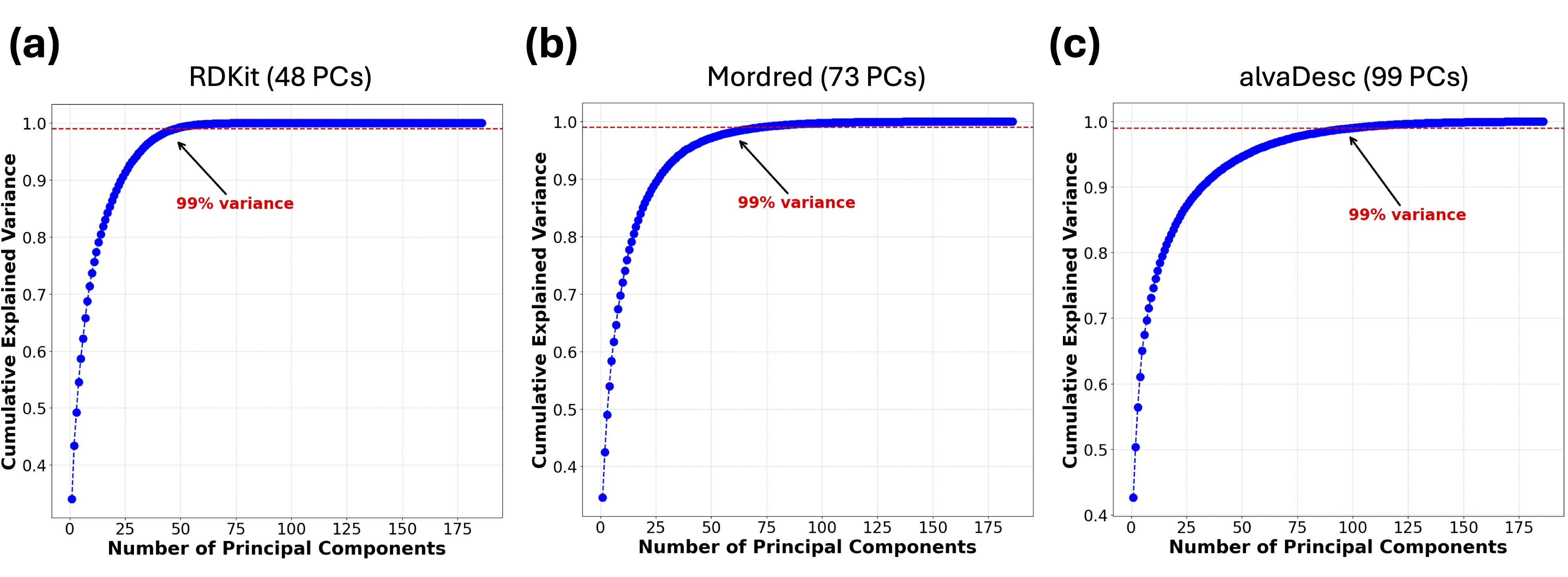}
\caption[]{Cumulative explained variance as a function of the number of PCs for each molecular descriptor package: (a) RDKit (48 PCs), (b) Mordred (73 PCs), and (c) alvaDesc (99 PCs). The red dashed line represents the 99\% variance threshold used to select the PCs for dimensionality reduction.}
\label{AMZyCCEBpP}
\end{figure}

\subsubsection{Neural Network Modeling Phase}

The PCs derived from each descriptor package (RDKit, Mordred, and AlvaDesc) were used as exploratory variables, while the quantile-transformed GWP 100 values served as outputs in their respective neural network models. The auto-tuning feature in Multi-Sigma was used to optimize the hyperparameters, including hidden layers, the number of neurons, activation functions, batch sizes, and epochs for each descriptor package. The top three models for each descriptor package, selected based on the lowest RMSE during validation, are presented in Table 1.

\begin{table}
\centering
\caption[]{Summary of the top three neural network models for each molecular descriptor package (RDKit, Mordred, and alvaDesc), including architecture details and performance metrics. QT refers to the quantile transform uniform scale, while Orig represents the original scale.}
\label{KFH9d7gHOO}
\begin{tabular}{p{\dimexpr 0.100\linewidth-2\tabcolsep}p{\dimexpr 0.100\linewidth-2\tabcolsep}p{\dimexpr 0.100\linewidth-2\tabcolsep}p{\dimexpr 0.100\linewidth-2\tabcolsep}p{\dimexpr 0.100\linewidth-2\tabcolsep}p{\dimexpr 0.100\linewidth-2\tabcolsep}p{\dimexpr 0.100\linewidth-2\tabcolsep}p{\dimexpr 0.100\linewidth-2\tabcolsep}p{\dimexpr 0.100\linewidth-2\tabcolsep}p{\dimexpr 0.100\linewidth-2\tabcolsep}}
\toprule
Parameter & \multicolumn{3}{p{\dimexpr 0.300\linewidth-2\tabcolsep}}{RDKit} & \multicolumn{3}{p{\dimexpr 0.300\linewidth-2\tabcolsep}}{Mordred} & \multicolumn{3}{p{\dimexpr 0.300\linewidth-2\tabcolsep}}{alvaDesc} \\
\hline
Best model & Model 1 & Model 2 & Model 3 & Model 1 & Model 2 & Model 3 & Model 1 & Model 2 & Model 3 \\
Epochs & 2674 & 6190 & 4895 & 4947 & 9877 & 1858 & 1001 & 5614 & 3669 \\
layers & 3 & 7 & 2 & 10 & 2 & 1 & 10 & 9 & 9 \\
Neurons & 82 & 95 & 50 & 91 & 2 & 29 & 21 & 59 & 69 \\
Batch size & 75 & 51 & 100 & 111 & 81 & 24 & 125 & 54 & 167 \\
Activation & tanh & sigmoid & sigmoid & sigmoid & sigmoid & sigmoid & sigmoid & sigmoid & sigmoid \\
RMSE (QT) & 0.0644 & 0.0677 & 0.0691 & 0.0581 & 0.0646 & 0.0674 & 0.0681 & 0.0715 & 0.0887 \\
R\textsuperscript{2} (QT) & 0.975 & 0.976 & 0.974 & 0.975 & 0.979 & 0.986 & 0.97 & 0.978 & 0.942 \\
RMSE (Orig) & 414.91 & 510.64 & 520.12 & 310.57 & 1497.38 & 1744.69 & 168.15 & 386.79 & 808.94 \\
R\textsuperscript{2} (Orig) & 0.96 & 0.91 & 0.98 & 1 & 0.83 & 0.88 & 1 & 0.98 & 0.93 \\
\bottomrule
\end{tabular}
\end{table}

The hyperparameter tuning results revealed substantial differences in the architectures and performance across models built from different descriptor packages. The RDKit-based models performed well with a simpler architecture, requiring between 2 and 7 hidden layers and a relatively lower number of neurons (between 50 and 95). Model 1 achieved an RMSE of 0.0644 on the quantile-transformed scale, an RMSE of 414.91, and an R\textsuperscript{2} score of 0.96 on the original scale. The Mordred-based models benefited from deeper architectures, with up to 10 hidden layers and 91 neurons per layer. This is likely because of the high feature complexity (73 PCs) derived from the descriptors, which include intricate 2D and 3D features. The Mordred models achieved the lowest RMSE, with model 1 achieving an RMSE of 0.0581 on the quantile-transformed scale, an RMSE of 310.57, and an R\textsuperscript{2} score of 1 on the original scale. AlvaDesc-based models utilized architectures with 9 to 10 hidden layers corresponding to the more detailed features captured (99 PCs after PCA). The models achieved RMSEs ranging from 0.0681 to 0.0887 on the quantile-transformed scale, with model 1 showing the best performance among alvaDesc-based models.

The tanh activation function used in RDKit Model 1 may have provided stability during training, while the sigmoid function was predominantly used in Mordred and AlvaDesc models, which likely helped capture complex relationships involving high-GWP refrigerants. However, deeper architectures with sigmoid functions could potentially lead to gradient vanishing, which was addressed by increasing the number of neurons and layers [18].

Using quantile transformation mitigated the positive skewness in the GWP data, enabling more efficient learning during model training. However, a trade-off was observed: while the transformation reduced RMSE on the quantile-transformed scale, predictions on the original GWP scale exhibited higher errors. This discrepancy suggests that normalization aids stability during training but does not fully address challenges posed by extreme values in the original data.

\subsubsection{Ensemble Model Testing Phase}

The top three models corresponding to each descriptor package were ensembled together to make predictions for the test set. Figure 4 shows the predictions made in the testing phase using the ensemble models, and individual model validation during the modeling phase.

\begin{figure}[!htbp]
\centering
\includegraphics[width=1\linewidth]{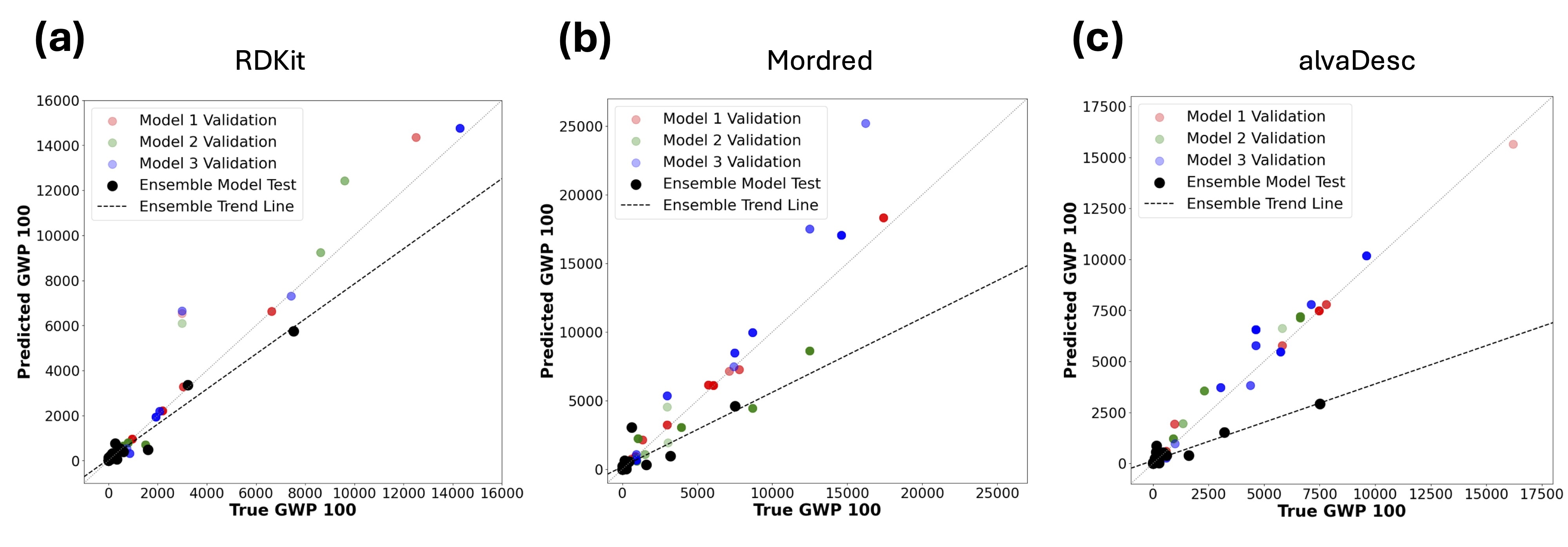}
\caption[]{Predicted vs. True GWP 100 values for the top three models and ensemble predictions for each molecular descriptor package. The dotted line represents the ideal trend, and the dashed line shows the ensemble trend}
\label{ZNPDEpmeTr}
\end{figure}

Among the three descriptor packages, the ensemble model utilizing RDKit demonstrated the best performance, with an RMSE of 481.9 and an R\textsuperscript{2} score of 0.918. This indicates high predictive accuracy and generalizability to the test dataset. In comparison, the ensemble models built on Mordred and alvaDesc descriptors had RMSE values of 1011.28 and 1118.78, respectively, with corresponding R\textsuperscript{2} scores of 0.641 for Mordred and 0.560 for alvaDesc. These results highlight that the RDKit-based ensemble outperformed the other models in terms of both predictive power and generalization capability.

The discrepancy in performance across descriptor packages suggests that the complexity of the descriptors plays a significant role in model generalization. The RDKit-based model, with simpler 2D descriptors, retained a robust predictive ability on the unseen test set, indicating its robustness against overfitting. In contrast, the Mordred and AlvaDesc descriptors, which include more intricate 2D and 3D features, appeared to introduce additional model complexity that hindered their ability to generalize effectively to new data. This trend is visually reflected in Figure 4, where the RDKit-based model shows predictions closer to the ideal line, suggesting a better fit across the entire range of GWP values.

The results of this study highlight that simpler molecular descriptors, like those from RDKit, may provide a better balance between model interpretability and generalizability for predicting GWP 100 values. The successful application of ensemble modeling also underscores the value of aggregating multiple models to enhance reliability, particularly in reducing the influence of overfitting, as seen with the Mordred and alvaDesc models.

Beyond GWP, the framework established here---leveraging dimensionality reduction, normalization, and ensemble deep learning---has significant potential for broader applications. Future research could adapt this methodology to predict other critical molecular properties, such as refrigerant toxicity or atmospheric lifetimes, enabling a more comprehensive and systematic screening process for designing sustainable refrigerants.

\subsection{Critical Molecular Descriptors and Their Role in GWP Reduction}

To identify the critical molecular descriptors influencing the GWP, factor analysis was conducted using the RDKit-based ensemble model due to its higher prediction accuracy. Figure 5 illustrates the ten most influential PCs on the GWP. Although all PCs were retained in the model to ensure comprehensive coverage (reaching 99\% variance explained), the focus here is on the top three contributing PCs---PC10, PC3, and PC4---based on their sensitivity and impact on the GWP 100. PC10 exhibited the most significant positive contribution to GWP predictions, indicating that features linked to PC10 are associated with increased GWP values. PC3 and PC4 demonstrated a significant negative impact, suggesting that features linked to PC3 and PC4 contribute to reducing GWP values.

\begin{figure}[!htbp]
\centering
\includegraphics[width=0.8\linewidth]{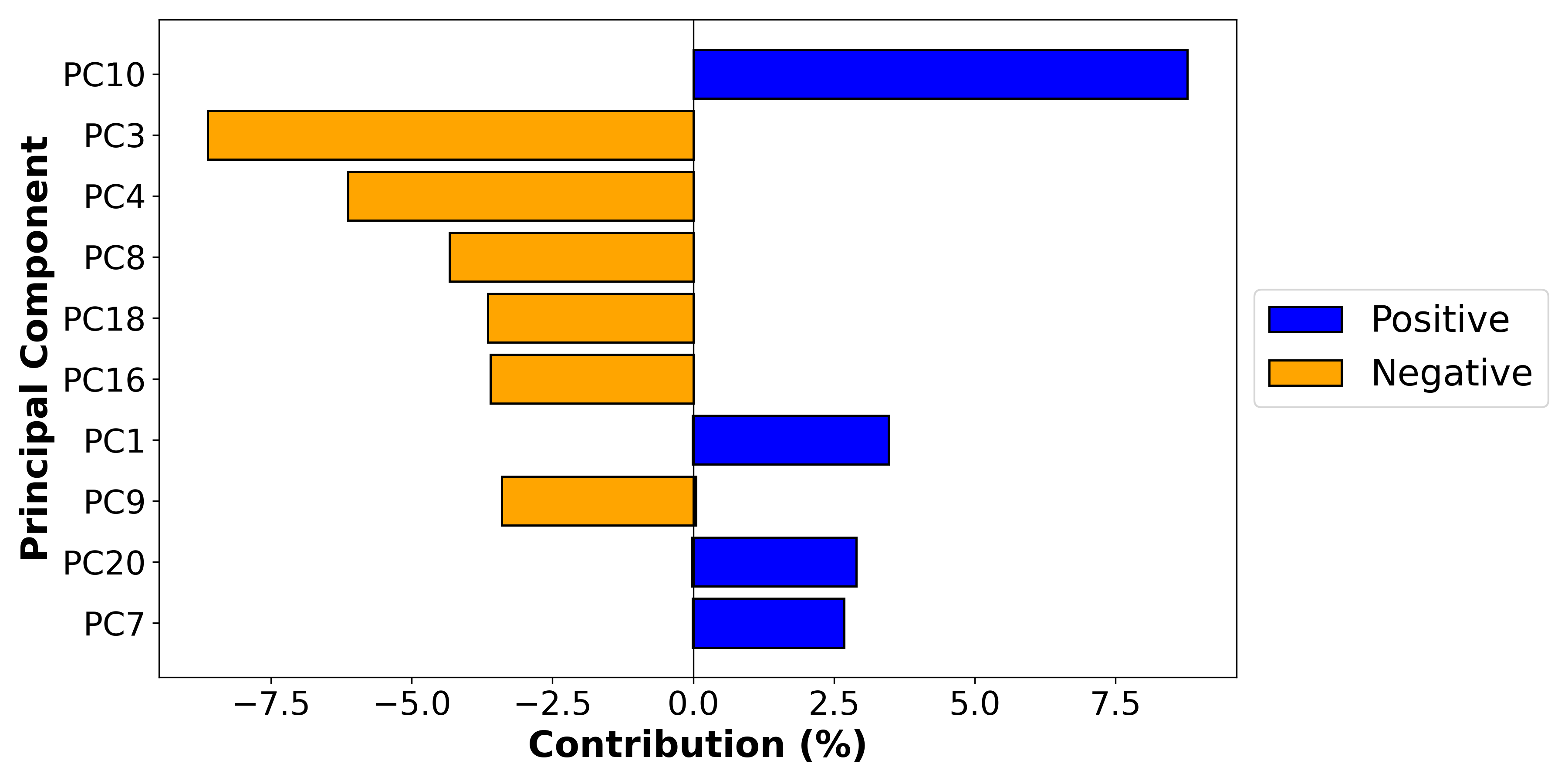}
\caption[]{Contribution of principal components (PCs) to GWP predictions based on factor analysis of RDKit-based ensemble model.}
\label{po7rxzojJ3}
\end{figure}

The key RDKit descriptors for each of these PCs were analyzed based on their loading values to interpret their chemical meaning (Table 2). Descriptors like BCUT2D\_MWLOW, SlogP\_VSA6, and fr\_allylic\_oxid were highly loaded on PC10. These descriptors relate to molecular weight distribution, lipophilicity, and the presence of a specific functional group (allylic oxide). PC10's positive impact suggests that higher molecular weight and specific functional groups correlate with increased GWP values. This likely reflects their chemical stability and long atmospheric lifetime [19].

\begin{table}
\centering
\caption[]{PC loadings for the most influential descriptors in predicting GWP}
\label{tYiFjapg1m}
\begin{tabular}{p{\dimexpr 0.333\linewidth-2\tabcolsep}p{\dimexpr 0.333\linewidth-2\tabcolsep}p{\dimexpr 0.333\linewidth-2\tabcolsep}}
\toprule
Principle component & Descriptor & Loading \\
\hline
PC10 & BCUT2D\_MWLOW & 0.2688 \\
PC10 & SlogP\_VSA6 & -0.2534 \\
PC10 & SMR\_VSA7 & 0.2594 \\
PC10 & fr\_allylic\_oxid & -0.2016 \\
PC10 & FpDensityMorgan3 & -0.196 \\
PC3 & SlogP\_VSA7 & 0.2382 \\
PC3 & Chi4v & 0.229 \\
PC3 & NumAliphaticHeterocycles & 0.2283 \\
PC3 & NumSaturatedHeterocycles & 0.2283 \\
PC3 & Chi2v & 0.1918 \\
PC4 & SMR\_VSA10 & -0.1934 \\
PC4 & SMR\_VSA9 & 0.1854 \\
PC4 & fr\_nitrite & 0.1854 \\
PC4 & SMR\_VSA2 & 0.1854 \\
PC4 & SlogP\_VSA4 & 0.1854 \\
\bottomrule
\end{tabular}
\end{table}

Important descriptors for PC3 included SlogP\_VSA7, Chi4v, and NumAliphaticHeterocycles. These relate to lipophilicity, molecular topological indices, and the presence of aliphatic heterocycles. PC3's negative impact suggests that these features potentially contribute to molecules' destabilization or reduced atmospheric persistence, thereby reducing GWP values.

The primary descriptors loaded on PC4 were SMR\_VSA10, fr\_nitrile, and SlogP\_VSA4, encompassing volume-related attributes and nitrile groups. We speculate that these features may enhance atmospheric reactivity under specific chemical conditions, promoting degradation and reducing GWP. Alternatively, they may stabilize the molecule in other contexts, prolonging its atmospheric lifetime and increasing GWP. Nitrile groups may make molecules less reactive with hydroxyl radicals (stabilizing effect) but more likely to persist under other atmospheric conditions [20]. Volume-related descriptors can influence how well molecules partition into reactive atmospheric phases, affecting degradation rates [21].

\section{CONCLUSION}

This study showcased the successful use of the Multi-Sigma platform in developing a fully connected neural network to predict the 100-year GWP of single-component refrigerants. Leveraging molecular descriptors and an advanced preprocessing framework, the ensembled neural network model captured complex structure-property relationships, demonstrating strong potential for virtual screening applications to identify environmentally friendly refrigerants.

Among the descriptor packages evaluated, the RDKit-based ensemble model achieved superior performance, with an RMSE of 481.9 and an R\textsuperscript{2} of 0.918 on the test set, indicating accuracy and generalizability. Dimensionality reduction using PCA ensured computational efficiency without significant information loss, while quantile transformation addressed skewness, leading to stable and reliable predictions.

Factor analysis revealed key molecular features influencing GWP, such as molecular weight distribution, lipophilicity, and functional groups like nitriles and allylic oxides. These insights provide actionable guidance for refrigerant design, emphasizing the importance of avoiding features linked to high GWP while enhancing those associated with environmental degradability. Additionally, the factor analysis highlights the value of interpretable machine learning models in deriving meaningful chemical insights.

While this study focused on single-component refrigerants, future work could explore the applicability of this methodology to complex refrigerant blends or related properties such as atmospheric lifetimes. The framework presented here, combining PCA, quantile transformation, and deep learning, is computationally efficient and scalable, making it suitable for large-scale virtual screening tasks. This study contributes directly to global climate mitigation efforts and the design of next-generation sustainable chemicals by accelerating the identification of low-GWP refrigerants.

\section{References}

[1] R. Llopis, D. Sánchez, R. Cabello, J. Catalán-Gil, and L. Nebot-Andrés, ``Experimental analysis of R-450A and R-513A as replacements of R-134a and R-507A in a medium temperature commercial refrigeration system,'' Int. J. Refrig., vol. 84, pp. 52--66, Dec. 2017, doi: 10.1016/j.ijrefrig.2017.08.022.

[2] H. Flerlage, G. J. M. Velders, and J. De Boer, ``A review of bottom-up and top-down emission estimates of hydrofluorocarbons (HFCs) in different parts of the world,'' Chemosphere, vol. 283, p. 131208, Nov. 2021, doi: 10.1016/j.chemosphere.2021.131208.

[3] P. Purohit et al., ``Electricity savings and greenhouse gas emission reductions from global phase-down of hydrofluorocarbons,'' Mar. 11, 2020. doi: 10.5194/acp-2020-193.

[4] G. J. M. Velders, D. W. Fahey, J. S. Daniel, S. O. Andersen, and M. McFarland, ``Future atmospheric abundances and climate forcings from scenarios of global and regional hydrofluorocarbon (HFC) emissions,'' Atmos. Environ., vol. 123, pp. 200--209, Dec. 2015, doi: 10.1016/j.atmosenv.2015.10.071.

[5] M. Kagzi, S. Khanra, and S. K. Paul, ``Machine learning for sustainable development: leveraging technology for a greener future,'' J. Syst. Inf. Technol., vol. 25, no. 4, pp. 440--479, Nov. 2023, doi: 10.1108/JSIT-11-2022-0266.

[6] I. I. I. Alkhatib, C. G. Albà, A. S. Darwish, F. Llovell, and L. F. Vega, ``Searching for Sustainable Refrigerants by Bridging Molecular Modeling with Machine Learning,'' Ind. Eng. Chem. Res., vol. 61, no. 21, pp. 7414--7429, Jun. 2022, doi: 10.1021/acs.iecr.2c00719.

[7] S. Devotta, A. Chelani, and A. Vonsild, ``Prediction of global warming potentials of refrigerants and related compounds from their molecular structure -- An artificial neural network with group contribution method,'' Int. J. Refrig., vol. 131, pp. 756--765, Nov. 2021, doi: 10.1016/j.ijrefrig.2021.08.011.

[8] R. Gani, ``Group contribution-based property estimation methods: advances and perspectives,'' Curr. Opin. Chem. Eng., vol. 23, pp. 184--196, Mar. 2019, doi: 10.1016/j.coche.2019.04.007.

[9] A. Sato, T. Miyao, S. Jasial, and K. Funatsu, ``Comparing predictive ability of QSAR/QSPR models using 2D and 3D molecular representations,'' J. Comput. Aided Mol. Des., vol. 35, no. 2, pp. 179--193, Feb. 2021, doi: 10.1007/s10822-020-00361-7.

[10] Y. Sun, X. Wang, N. Ren, Y. Liu, and S. You, ``Improved Machine Learning Models by Data Processing for Predicting Life-Cycle Environmental Impacts of Chemicals,'' Environ. Sci. Technol., vol. 57, no. 8, pp. 3434--3444, Feb. 2023, doi: 10.1021/acs.est.2c04945.

[11] Y. Xue, Z. R. Li, C. W. Yap, L. Z. Sun, X. Chen, and Y. Z. Chen, ``Effect of Molecular Descriptor Feature Selection in Support Vector Machine Classification of Pharmacokinetic and Toxicological Properties of Chemical Agents,'' J. Chem. Inf. Comput. Sci., vol. 44, no. 5, pp. 1630--1638, Sep. 2004, doi: 10.1021/ci049869h.

[12] G. Zhao, H. Kim, C. Yang, and Y. G. Chung, ``Leveraging Machine Learning To Predict the Atmospheric Lifetime and the Global Warming Potential of SF 6 Replacement Gases,'' J. Phys. Chem. A, vol. 128, no. 12, pp. 2399--2408, Mar. 2024, doi: 10.1021/acs.jpca.3c07339.

[13] Intergovernmental Panel on Climate Change (IPCC), Climate Change 2021 -- The Physical Science Basis: Working Group I Contribution to the Sixth Assessment Report of the Intergovernmental Panel on Climate Change. Cambridge: Cambridge University Press, 2023. doi: 10.1017/9781009157896.

[14] F. P. Duchesnay Gaël Varoquaux, Alexandre Gramfort, Vincent Michel, Bertrand Thirion, Olivier Grisel, Mathieu Blondel, Peter Prettenhofer, Ron Weiss, Vincent Dubourg, Jake Vanderplas, Alexandre Passos, David Cournapeau, Matthieu Brucher, Matthieu Perrot, Édouard, ``Scikit-learn: Machine Learning in Python,'' J. Mach. Learn. Res., vol. 12, pp. 2825--2830, 2011.

[15] B. Casier, S. Carniato, T. Miteva, N. Capron, and N. Sisourat, ``Using principal component analysis for neural network high-dimensional potential energy surface,'' J. Chem. Phys., vol. 152, no. 23, p. 234103, Jun. 2020, doi: 10.1063/5.0009264.

[16] L.-Y. Wu and S.-S. Weng, ``Ensemble Learning Models for Food Safety Risk Prediction,'' Sustainability, vol. 13, no. 21, p. 12291, Nov. 2021, doi: 10.3390/su132112291.

[17] M. Shahhosseini, G. Hu, and S. V. Archontoulis, ``Forecasting Corn Yield With Machine Learning Ensembles,'' Front. Plant Sci., vol. 11, p. 1120, Jul. 2020, doi: 10.3389/fpls.2020.01120.

[18] M. Roodschild, J. Gotay Sardiñas, and A. Will, ``A new approach for the vanishing gradient problem on sigmoid activation,'' Prog. Artif. Intell., vol. 9, no. 4, pp. 351--360, Dec. 2020, doi: 10.1007/s13748-020-00218-y.

[19] C. M. Roehl, D. Boglu, C. Brühl, and G. K. Moortgat, ``Infrared band intensities and global warming potentials of CF4 , C2 F6 , C3 F8 , C4 F10 , C5 F12 , and C6 F14,'' Geophys. Res. Lett., vol. 22, no. 7, pp. 815--818, Apr. 1995, doi: 10.1029/95GL00488.

[20] J. Hioe, D. Šakić, V. Vrček, and H. Zipse, ``The stability of nitrogen-centered radicals,'' Org. Biomol. Chem., vol. 13, no. 1, pp. 157--169, 2015, doi: 10.1039/C4OB01656D.

[21] R. Koppmann, ``Chemistry of Volatile Organic Compounds in the Atmosphere,'' in Hydrocarbons, Oils and Lipids: Diversity, Origin, Chemistry and Fate, H. Wilkes, Ed., Cham: Springer International Publishing, 2020, pp. 811--822. doi: 10.1007/978-3-319-90569-3\_24.

\section*{Acknowledgments}
The authors gratefully acknowledge the financial support provided by the New Energy and Industrial Technology Development Organization (NEDO) under the project JPNP23001. This research would not have been possible without their generous funding and commitment to advancing sustainable technologies.

\end{document}